\documentclass[lettersize,journal]{IEEEtran}
\usepackage{amsmath,amsfonts}
\usepackage{algorithmic}
\usepackage{algorithm}
\usepackage{array}
\usepackage[caption=false,font=normalsize,labelfont=sf,textfont=sf]{subfig}
\usepackage{textcomp}
\usepackage{stfloats}
\usepackage{url}
\usepackage{verbatim}
\usepackage{graphicx}
\usepackage{cite}
\usepackage{booktabs} 
\usepackage{multirow}
\usepackage{graphicx}
\usepackage[table]{xcolor}

\begin{document}

\bstctlcite{IEEEexample:BSTcontrol}
\title{SG-Splat: Stable Geometry Learning under Pose Uncertainty for Pose-Free 3D Gaussian Splatting}


\author{
	Xiaohan~Lu,
	Jiaqi~Zhang,
        Weijia~Jiang,
	Siwei~Ma, ~\IEEEmembership{Fellow,~IEEE}	
		\thanks{
		}
		\thanks{Xiaohan~Lu, Jiaqi~Zhang, and Siwei~Ma are with the State Key Laboratory of Multimedia Information Processing, School of Computer Science, Peking University, Beijing 100871, China~(e-mail: luxiaohan@stu.pku.edu.cn, jqzhang@pku.edu.cn, swma@pku.edu.cn). }
        \thanks{Weijia~Jiang is with the Migu Interactive Entertainment Co., Ltd, Nanjing 210000, China~(email: jiangweijia@migu.chinamobile.com).}
	}

\markboth{Journal of \LaTeX\ Class Files,~Vol.~14, No.~8, August~2021}%
{Shell \MakeLowercase{\textit{et al.}}: A Sample Article Using IEEEtran.cls for IEEE Journals}


\maketitle

\begin{abstract}
3D Gaussian Splatting (3DGS) has become a promising representation for real-time novel view synthesis and 3D scene reconstruction. 
For practical deployment, recent studies have explored pose-free feed-forward 3DGS, which reconstructs Gaussian representations directly from unposed input images. 
However, these methods remain sensitive to residual pose errors, which perturb canonical-space Gaussian transformation, introducing noisy supervision for geometry and attribute learning. 
To address this problem, we present SG-Splat, a pose-free feed-forward 3DGS framework that improves robustness to pose uncertainty and stabilizes  the learning of Gaussian primitives. 
SG-Splat decouples geometry estimation and Gaussian attribute prediction into two dedicated branches, reducing error coupling between pose-sensitive geometry and Gaussian attributes.
We further introduce a coarse-to-fine refinement pipeline, consisting of Flow-Guided Pose Refinement (FGPR) and Residual-Guided Geometry Correction (RGC), to improve geometric alignment before Gaussian attribute reconstruction. 
Specifically, FGPR refines camera poses with flow-based correlation features and differentiable bundle adjustment. 
RGC then corrects initial point maps using depth residuals derived from FGPR, producing more reliable Gaussian centers. 
To stabilize training, we introduce a dual-path strategy that mixes ground-truth and predicted poses for canonical transformation. 
It provides clean supervision for Gaussian attributes while improving geometry robustness to residual pose errors.
Extensive experiments on DL3DV and RealEstate10K show that SG-Splat achieves state-of-the-art performance in both novel view synthesis and pose estimation. 
\end{abstract}


\section{Introduction}
\begin{figure}[t]
\centering
\includegraphics[width=\linewidth, trim=0cm 15.5cm 8cm 0cm, clip]{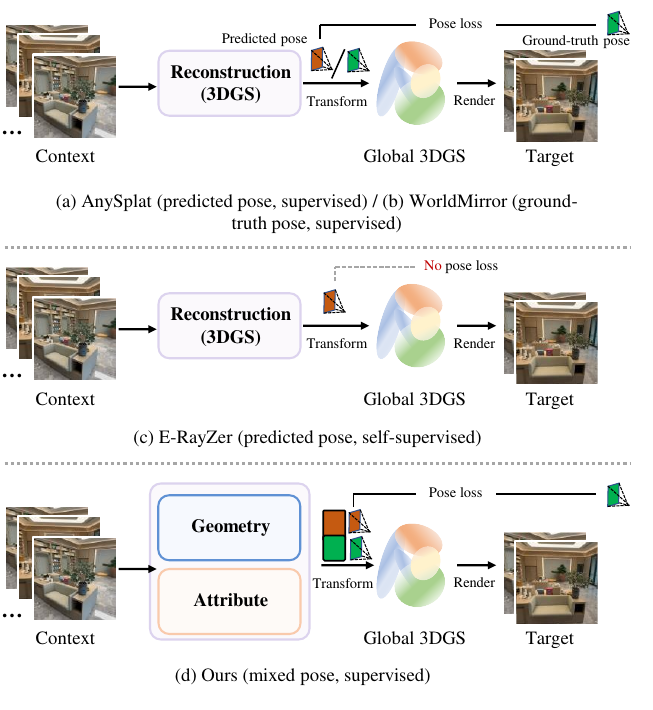}
\caption{Comparison of pose usage strategies in pose-free feed-forward 3DGS training. 
(a) AnySplat~\cite{jiang2025anysplat} uses predicted poses with pose supervision, maintaining consistency between training and inference but suffering from perturbations caused by pose errors.
(b) WorldMirror~\cite{liu2025worldmirror} uses ground-truth poses with pose supervision, stabilizing training but introducing inconsistency between training and inference. 
(c) E-RayZer\cite{zhao2025rayzer} relies on predicted poses without pose supervision, making optimization sensitive to pose errors. 
(d) Our method uses mixed poses with pose supervision, balancing stable learning and robustness to residual pose errors.}
\label{fig:pose_sel}
\end{figure}
3D Gaussian Splatting (3DGS)\cite{kerbl20233d3dgs} has become a powerful representation for real-time photorealistic rendering and high-fidelity scene reconstruction, with growing potential in world modeling~\cite{worldmodel_survey}, autonomous driving~\cite{3dgs_review}, and virtual reality~\cite{song2026tvcg_3dgsvr}. However, conventional 3DGS is limited by its reliance on per-scene optimization and calibrated poses, which limits scalable and efficient 3D reconstruction.
This has motivated recent interest in pose-free feed-forward 3DGS, which seeks to predict camera poses and Gaussian parameters directly from input images in a single forward pass. 
Early pose-free feed-forward 3DGS methods~\cite{noposplat,smart2024splatt3r} predict 3D Gaussian primitives directly in a canonical space, thereby avoiding the explicit propagation of pose estimation errors during Gaussian aggregation. These methods have shown promising performance under sparse-view settings, typically with two to four input views. However, directly constructing a canonical Gaussian representation becomes increasingly challenging as the number of input views grows\cite{zhang2025flare}, due to the difficulty of maintaining cross-view geometric consistency without explicit pose-guided aggregation.

To scale to multi-view settings, recent methods such as AnySplat~\cite{jiang2025anysplat} and WorldMirror~\cite{liu2025worldmirror} predict per-view Gaussian primitives in camera coordinate systems and transform them into a canonical space using camera poses of each view, as illustrated in Fig.~\ref{fig:pose_sel}(a) and (b). At inference time, this transformation must rely on predicted poses, whose errors directly affect the alignment of Gaussian primitives across views. During training, this transformation becomes non-trivial, as the poses used for transformation affect both optimization stability and the consistency between training and inference. AnySplat uses predicted poses for transformation during training, improving consistency with inference but introducing scene-dependent pose perturbations that destabilize the learning of geometry and Gaussian attributes. In contrast, WorldMirror uses ground-truth poses for canonical transformation, which stabilizes Gaussian parameter learning but introduces a discrepancy between training and inference. 
In addition, both methods directly couple pose supervision with Gaussian reconstruction, introducing competing optimization objectives. This coupling becomes especially detrimental when residual pose errors perturb the canonical transformation used for rendering supervision.
To remove this coupling, recent methods~\cite{huang2025spfsplatv2,zhao2025rayzer} adopt a self-supervised formulation.
As shown in Fig.~\ref{fig:pose_sel}(c), they discard explicit pose supervision and optimize Gaussian reconstruction solely through rendering supervision. 
Although this design removes the need for pose supervision, it still relies on predicted poses for canonical transformation during training. Consequently, pose errors still perturb the optimization of Gaussian geometry and attributes. In addition, without direct 3D constraints, these methods may produce visually plausible but geometrically unreliable reconstructions~\cite{mitchelryu2025xfactor}, restricting their applicability in geometry-sensitive tasks.

These observations reveal a key challenge: \textbf{how can pose-free feed-forward 3DGS effectively exploit pose supervision during training while maintaining stable learning under residual pose errors?}
Although YoNoSplat~\cite{ye2025yonosplat} mixes predicted and ground-truth poses during training, predicted poses are sampled with a low probability. 
As a result, this strategy only weakly exposes the model to inference-time pose errors and still allows noisy supervision to affect Gaussian attributes when predicted poses are selected.

To address this challenge, we propose SG-Splat, a pose-free feed-forward 3DGS framework with decoupled geometry and attribute branches, as shown in Fig.~\ref{fig:pose_sel}(d). 
This design separates geometry and camera estimation from Gaussian attribute prediction, reducing the interference between pose errors and attribute learning. 
We further introduce a progressive refinement pipeline with Flow-Guided Pose Refinement (FGPR) and Residual-Guided Geometry Correction (RGC). 
FGPR refines the initial camera poses using flow-based correlation features and differentiable bundle adjustment (DBA). 
RGC corrects the initial point maps using depth residuals produced by FGPR and intermediate geometry features. 
The refined poses stabilize canonical transformations, and the corrected point maps improve Gaussian center accuracy.
Together, they reduce the discrepancy between training and inference and improve the consistency between the refined camera poses and the Gaussian centers.

In addition, we design a pose-aware dual-path training strategy that prevents the attribute branch from overfitting to alignments corrupted by residual pose errors.
Specifically, either ground-truth poses or predicted poses are selected to transform Gaussian primitives into a canonical coordinate system during training. 
When ground-truth poses are used, the model learns geometry reconstruction and Gaussian attribute prediction under reliable alignment. 
When predicted poses are used, we mainly optimize the geometry branch and pose estimation modules, preventing the attribute branch from overfitting to residual pose errors. 
This asymmetric optimization stabilizes Gaussian attribute learning and improves geometry robustness under pose uncertainty.
Our contributions are summarized as follows:
\begin{itemize}
    \item We propose SG-Splat, a pose-free feed-forward 3DGS framework with a geometry-attribute decoupled architecture. By separating geometry and camera estimation from Gaussian attribute prediction, the framework suppresses error propagation from inaccurate pose estimates and improves the reliability of Gaussian parameter generation.
    \item We introduce a coarse-to-fine refinement pipeline consisting of FGPR and RGC to enhance pose and Gaussian alignment. 
    FGPR iteratively refines camera poses by combining flow-based correlation features with differentiable bundle adjustment. 
    RGC then corrects local point maps using depth residuals from FGPR, improving the consistency between refined camera poses and Gaussian centers.
    \item We develop a pose-aware dual-path training strategy that integrates mixed-pose canonical transformation with asymmetric branch optimization. This strategy uses ground-truth poses to provide reliable attribute supervision, while using predicted poses to regularize geometry learning against scene-dependent pose errors.
\end{itemize}

\section{Related Work}

\subsection{Novel View Synthesis with Feed-Forward 3D Gaussian Splatting}

Early image-based rendering methods~\cite{levoy2023light,lumigraph,buehler2001unstructured} synthesize novel views by resampling or blending images captured from nearby            viewpoints. These methods demonstrated the possibility of photorealistic view synthesis without fully explicit 3D reconstruction, but they often require dense image capture, calibrated cameras, or approximate geometry to handle large viewpoint changes and occlusions.
With the advent of deep learning, learning-based NVS methods~\cite{flynn2016deepstereonvs,kalantari2016learningnvs}learn the geometry of scenes and appearance priors from data, improving flexibility under sparse-view inputs. 
NeRF~\cite{mildenhall2021nerf} represents a scene as a continuous radiance field optimized from posed images and renders novel views via volumetric ray marching, which incurs substantial computational cost. To improve efficiency, follow-up works~\cite{muller2022instant, fridovich2022plenoxels, chen2022tensorf} introduced compact encodings or hybrid scene representations to accelerate rendering, but they still struggle to achieve real-time performance. Motivated by this limitation, 3DGS~\cite{kerbl20233d3dgs} and its variants~\cite{jiang2026tvcg_geote,zhou2026tvcg_sggs,cai2024tvcg_dust} adopt explicit Gaussian primitives for scene representation, achieving high-fidelity novel view synthesis with real-time rendering efficiency.
However, both NeRF and 3DGS pipelines typically rely on per-scene optimization. 
Recent methods extend 3DGS to sparse input views by introducing diffusion priors~\cite{liu2026tip_reconx} or geometric constraints~\cite{bao2025tip_loopsparsegs}. 
This has motivated recent feed-forward methods for NeRF~\cite{yu2021pixelnerf,chen2021mvsnerf,chen2025tvcg_volume} as well as feed-forward 3DGS methods~\cite{pixelsplat,bao2026tvcg_fisn,zhang2024gs-lrm,song2026tinysplat}. More recently, DepthSplat~\cite{xu2024depthsplat} and ProSplat~\cite{lu2026prosplat} extend this paradigm to sparse multi-view or wide-baseline settings. In addition, recent works~\cite{moreau2025off,kang2025ilrm,ren2026tokengs} decouple Gaussian primitives from pixel-aligned locations, resulting in a more flexible representation.

Nevertheless, most of them still assume known camera poses or rely on pose-conditioned geometric cues, which limits their applicability to casually captured, unposed image collections. In contrast, our work targets pose-free feed-forward 3DGS, where camera poses, geometry, and Gaussian attributes must be jointly predicted under residual pose uncertainty.
\subsection{3D Foundation Models for Geometry and Pose Priors}
Recent 3D foundation models have substantially advanced feed-forward geometry and pose estimation from unposed images. DUSt3R~\cite{wang2024dust3r} reformulates multi-view reconstruction as point map regression from unconstrained image pairs. MASt3R~\cite{mast3r} and its variants\cite{yang2025fast3r,chen2026lingbomap} further improve dense matching accuracy and geometric consistency. Building on this line of work, VGGT~\cite{wang2025vggt}, $\pi^3$~\cite{wang2025pi3}, and Depth Anything 3 (DA3)~\cite{lin2025depthanything3} directly predict camera poses, point maps, depth maps, and geometric tracks from a few to hundreds of views in a single forward pass, offering scalable geometry priors for downstream applications.

These foundation models have also been increasingly adopted in feed-forward 3DGS. Splatt3R~\cite{smart2024splatt3r} and NoPoSplat~\cite{noposplat} build upon MASt3R to obtain geometric priors and predict Gaussians in a global coordinate system from unposed images. AnySplat~\cite{jiang2025anysplat} uses VGGT as a teacher to supervise pose and geometry prediction during training. YoNoSplat~\cite{ye2025yonosplat} extends this paradigm with a $\pi^3$-based architecture and introduces an intrinsic condition embedding to handle unknown camera intrinsics. AirSplat~\cite{bui2026airsplat} leverages DA3 as a frozen geometry foundation model and trains a Gaussian generation module on top of its priors. Collectively, these methods demonstrate that 3D foundation models have become a powerful source of geometric priors for pose-free feed-forward 3DGS.
\begin{figure*}[t]
\centering
\includegraphics[width=\linewidth, trim=0cm 22.2cm 0cm 0cm, clip]{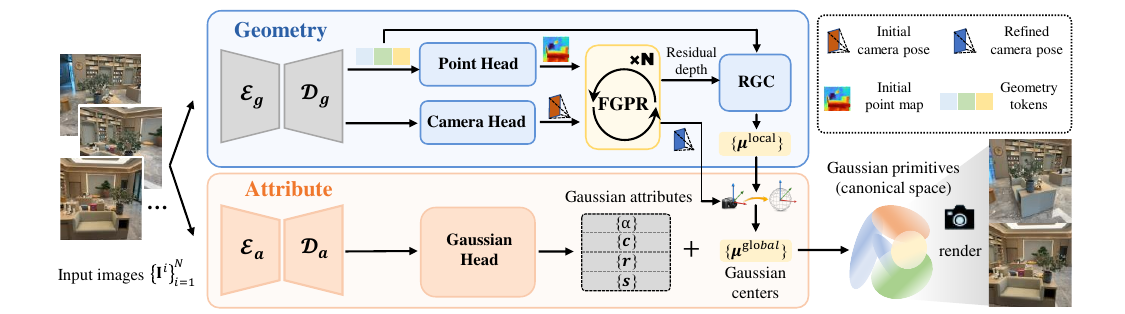}
\caption{\textbf{Overview of the pipeline.} 
The input images are processed by separate geometry and attribute branches. 
The geometry branch predicts initial local point maps and camera poses through the point head and camera head, while the attribute branch predicts the remaining Gaussian attributes. 
Flow-Guided Pose Refinement (FGPR) refines the initial poses and produces depth residuals, which are used by Residual-Guided Geometry Correction (RGC) to correct the local Gaussian centers $\boldsymbol{\mu}^{\mathrm{local}}$ . 
The corrected centers are transformed into the canonical space with the refined poses and combined with the predicted attributes to form the final Gaussian representation.
}
\label{fig:pipline}
\end{figure*}
\subsection{Pose-Free Feed-Forward 3D Gaussian Splatting}

Pose-free feed-forward 3DGS reconstructs Gaussian representations directly from unposed images, avoiding camera calibration at inference time. Existing methods mainly fall into two categories. The first line exploits 3D foundation models to provide geometry and camera priors. Methods built on DUSt3R~\cite{xu2025freesplatter} or MASt3R~\cite{smart2024splatt3r, noposplat, li2025vicasplat,zhang2025flare,li2026tip_scsv} recover geometry and camera poses before Gaussian prediction. However, these methods are mostly tailored to sparse input views, which limits their scalability to denser multi-view reconstruction.
Recent methods~\cite{jiang2025anysplat,ye2025yonosplat,bui2026airsplat,li2026tokensplat} improve scalability by using visual geometry foundation models that jointly predict point maps and camera poses in a single forward pass~\cite{wang2025vggt,wang2025pi3,lin2025depthanything3}. However, their Gaussian reconstruction remains sensitive to pose accuracy, because residual pose errors can perturb the transformation of Gaussian centers and introduce noisy supervision into the optimization of Gaussian geometry and attributes.

Another line of work removes 3D supervision and trains pose-free models with self-supervised rendering objectives~\cite{jiang2025rayzer,kang2025selfsplat,huang2025spfsplatv2,zhao2025rayzer}. Although these methods avoid explicit pose supervision, pose estimation, geometry prediction, and Gaussian attribute generation are optimized in a coupled manner through rendering losses. Consequently, pose and geometry errors may be absorbed by Gaussian attributes, leading to visually plausible but geometrically unreliable reconstructions~\cite{mitchelryu2025xfactor}.

In contrast, our work focuses on the practical setting where explicit 3D supervision is available. We study how to effectively exploit 3D foundation priors for dense-view pose-free feed-forward 3DGS while preventing pose-corrupted supervision from destabilizing Gaussian geometry and attribute learning.

\section{Method}

\subsection{Problem Formulation}
Given $N$ unposed input views
\(
\mathcal{I}=\{\mathbf{I}_i\}_{i=1}^{N},
\)
where $\mathbf{I}_i \in \mathbb{R}^{H \times W \times 3}$, the goal is to reconstruct a set of 3D Gaussian primitives for novel view synthesis. Each primitive is defined as
\begin{equation}
\mathbf{g}_{i,j}=
(\boldsymbol{\mu}_{i,j},\, \mathbf{c}_{i,j},\, \mathbf{r}_{i,j},\, \mathbf{s}_{i,j},\, \alpha_{i,j}),
\end{equation}
and the complete Gaussian representation is written as
\begin{equation}
\mathcal{G}=\{\mathbf{g}_{i,j}\}_{i=1,\ldots,N;\,j=1,\ldots,HW}.
\end{equation}
The components $\boldsymbol{\mu}\in\mathbb{R}^{3}$, $\mathbf{r}\in\mathbb{R}^{4}$, $\mathbf{s}\in\mathbb{R}^{3}$, $\alpha\in\mathbb{R}_{+}$, and $\mathbf{c}\in\mathbb{R}^{3\times (k+1)^2}$ correspond to the 3D position, quaternion rotation, scaling factor, opacity, and spherical harmonics (SH) coefficients of degree $k$, respectively.
We assume known camera intrinsics, which are directly obtained from the capture device.
In addition, the model predicts the camera poses $\{\mathbf{P}_i\}_{i=1}^{N}$ for all input views.
Formally, the feed-forward model $f_{\theta}$ parameterized by $\theta$ learns the mapping
\begin{equation}
f_{\theta}:\{\mathbf{I}_i\}_{i=1}^{N}
\mapsto
\left(
\mathcal{G},\,
\{\mathbf{P}_i\}_{i=1}^{N}
\right).
\end{equation}


\subsection{Pipeline}
\begin{figure*}[t]
\centering
\includegraphics[width=\linewidth, trim=0cm 23.0cm 0cm 0cm, clip]{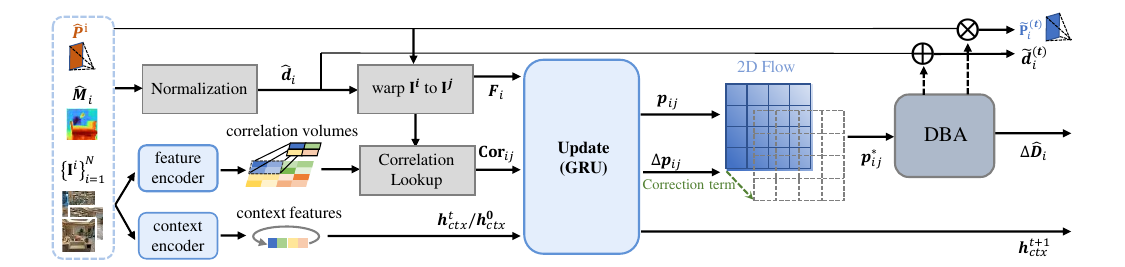}
\caption{\textbf{Flow-Guided Pose Refinement (FGPR).} At each iteration, FGPR uses the current pose and geometry estimates to build flow-based correlation features and optical flow of camera motion. 
Together with context features, these cues are fed into an update operator to predict correspondence corrections $\Delta \mathbf{p}_{i j}$. 
The corrected correspondences $\mathbf{p}^{*}_{i j}$ are optimized by differentiable bundle adjustment (DBA) to jointly update camera poses and depths. 
Over $T$ iterations, FGPR progressively improves pose accuracy and produces depth residuals for subsequent geometry correction.}
\label{fig:pose_refine}
\end{figure*}
As illustrated in Fig.~\ref{fig:pipline}, SG-Splat decouples geometry and attribute modeling into two independent branches, each consisting of an encoder and a decoder. Given the input images, the two branches produce a geometry token $\mathbf{z}_i^g$, a camera token $\mathbf{z}_i^c$, and an attribute token $\mathbf{z}_i^a$ for the $i$-th input view. The geometry token $\mathbf{z}_i^{g}$ and attribute token $\mathbf{z}_i^{a}$ are both of size $\mathbb{R}^{(l+5)\times d}$, while the camera token $\mathbf{z}_i^{c}$ is of size $\mathbb{R}^{(l/4+5)\times d/4}$, where $l=\frac{4HW}{p^2}$ denotes the number of image tokens, and the additional five tokens are register tokens following $\pi^3$~\cite{wang2025pi3}. Here, $p=14$ is the patch size and $d=1024$ is the token dimension.

The geometry token and the camera token are then fed into a point head and a camera head to predict an initial local point map $\hat{\mathbf{M}}_i^{local}$ and an initial camera pose $\hat{\mathbf{P}}_i$ for each input view, respectively. The predicted pose is further refined by the proposed Flow-Guided Pose Refinement (FGPR) module, which iteratively updates the pose for $T$ rounds and produces the refined pose $\tilde{\mathbf{P}}_i$ together with depth residual cues for geometric correction. The initial local point map $\hat{\mathbf{M}}_i^{local}$ is then refined by the Residual-Guided Geometry Correction module to obtain the refined local point map $\tilde{\mathbf{M}}_i^{local}$. Subsequently, $\tilde{\mathbf{M}}_i^{local}$ is transformed from the local camera coordinate system into the shared canonical space, yielding the global Gaussian centers $\{\boldsymbol{\mu}_{i,j}^{global}\}_{j=1}^{HW}$.

In parallel, the attribute token is fed into a Gaussian head to predict the remaining Gaussian attributes, which are combined with the transformed Gaussian centers to form the complete Gaussian representation:
\begin{equation}
\mathcal{G}=
\left\{
\left(
\boldsymbol{\mu}_{i,j}^{global},
\mathbf{c}_{i,j},
\mathbf{r}_{i,j},
\mathbf{s}_{i,j},
\alpha_{i,j}
\right)
\right\}_{i=1,\ldots,N;\,j=1,\ldots,HW}.
\end{equation}

\subsection{Flow-Guided Pose Refinement (FGPR)}
Inspired by DROID-SLAM\cite{teed2021droid}, we adopt an iterative pose refinement module to further correct the initial camera parameters $\hat{\mathbf{P}}_i$, as shown in Fig.~\ref{fig:point refine}. Given the input images $
\mathbf{I}_i
$, initial camera parameters$\hat{\mathbf{P}}_i$, and initial depth estimates $\hat{\mathbf{D}}_i$ from the z-axis $\hat{\mathbf{M}}_i$, the module constructs low-based correlation features and optical flow of camera motion to refine poses iteratively.

\paragraph{Inverse Depth Normalization}
To improve robustness to scene-scale variation, we use normalized inverse depth. First, we convert depth to inverse depth:
\begin{equation}
\hat{\mathbf{d}}_i^{raw} = \frac{1}{\hat{\mathbf{D}}_i}, \qquad
\hat{\mathbf{d}}_i^{raw} \in \mathbb{R}^{H' \times W'},
\end{equation}
where $H' = H/8$ and $W' = W/8$. We then normalize the inverse depth using only valid values,
\begin{equation}
\hat{\mathbf{d}}_i=\frac{\hat{\mathbf{d}}_i^{raw}}{\mathrm{mean}\bigl(\{\hat{\mathbf{d}}_i^{raw}[\hat{\mathbf{d}}_i^{raw}>\tau]\}_{i=1}^{N}\bigr)},
\end{equation}
where $\tau=0.05$ filters out extremely small inverse-depth values corresponding to very distant points. This normalization maintains a consistent numerical scale for the geometric input, thereby improving the robustness of pose refinement.

\paragraph{Feature Extraction}
For feature extraction, input images are processed by a feature encoder and a context encoder, producing two dense feature maps at $1/8$ image resolution. 
The feature encoder is used to compute 4D correlation
volumes\cite{teed2020raft} $C_{ij} \in \mathbb{R}^{H' \times W' \times H' \times W'}$ for each pair of overlapping images $(i,j)$.
Then, a correlation lookup operator takes a $H' \times W'$ grid of query coordinates and retrieves the values within a local square neighborhood of radius $r$ from $C_{ij}$, producing
$\textbf{Cor}_{ij} \in \mathbb{R}^{H' \times W' \times (2r+1)^2}$.
The context features are injected into the update network during iterative refinement.
\paragraph{Input}
Under the current pose estimates $\hat{\mathbf{P}}_i$ and $\hat{\mathbf{P}}_j$ and the inverse depth $\hat{\mathbf{d}}_i$, we warp the pixel grid $\mathbf{p}_i$ from view $i$ to view $j$ to obtain the correspondence field:
\begin{equation}
\mathbf{p}_{ij} = \Pi \!\left( {\hat{\mathbf{P}}_j}^{-1} \circ {\hat{\mathbf{P}}_i} \circ \Pi^{-1}(\mathbf{p}_i, \hat{\mathbf{d}}_i) \right),
\end{equation}
where $\mathbf{p}_i$ is the pixel grid of view $i$, and $\Pi^{-1}$ and $\Pi$ denote back-projection and projection, respectively. Based on the correspondence field $\mathbf{p}_i$, we derive the optical flow of camera motion $\mathbf{F}_i$ as the displacement between the projected and original pixel locations. Consequently, the update module takes the context feature $\mathbf{h}_{\mathrm{ctx}}$, the correlation feature $\mathbf{Cor}_{ij}$, and the optical flow $\mathbf{F}_i$ as inputs.

\paragraph{Update}
The update operator is implemented as a $3 \times 3$ convolutional GRU initialized with the context feature $\mathbf{h}_{\mathrm{ctx}}$. To improve feature fusion, the correlation feature $\mathbf{Cor}_{ij}$ and the optical flow $\mathbf{F}_i$ are first encoded by two convolutional layers before being injected into the GRU, while the context feature $\mathbf{h}_{\mathrm{ctx}}$ is incorporated through element-wise addition.

In iteration $t$, the operator updates the hidden state and predicts a correction term $\Delta \mathbf{p}_{ij}^{(t)}$ for the current correspondence field $\mathbf{p}_{ij}^{(t)}$:
\begin{equation}
\Delta \mathbf{p}_{ij}^{(t)} =
\mathrm{Update}\!\left(
\mathbf{h}_{\mathrm{ctx}},\,
\mathbf{Cor}_{ij}^{(t)},\,
\mathbf{F}_i^{(t)}
\right),
\end{equation}
\begin{equation}
\mathbf{p}_{ij}^{*(t)} =
\mathbf{p}_{ij}^{(t)} + \Delta \mathbf{p}_{ij}^{(t)}.
\end{equation}
The corrected correspondence field is then fed into the DBA~\cite{tang2018dba} module together with the current pose and inverse depth estimates $\tilde{\mathbf{P}}_i^{(t)}$ and $\tilde{\mathbf{d}}_i^{(t)}$, to jointly update them as $\tilde{\mathbf{P}}_i^{(t+1)}$ and $\tilde{\mathbf{d}}_i^{(t+1)}$.
After $T$ iterations, the final refined pose and depth are given by
\begin{equation}
\tilde{\mathbf{P}}_i = \tilde{\mathbf{P}}_i^{(T)}, \qquad
\tilde{\mathbf{D}}_i = \frac{1}{\tilde{\mathbf{d}}_i^{(T)}}.
\end{equation}
The depth residual used for subsequent geometry correction is defined as
\begin{equation}
\Delta \hat{\mathbf{D}}_i = \tilde{\mathbf{D}}_i - \hat{\mathbf{D}}_i,
\end{equation}
where $\hat{\mathbf{D}}_i$ denotes the initial depth estimate.

\subsection{Residual-Guided Geometry Correction (RGC).}
\begin{figure}[t]
\centering
\includegraphics[width=\linewidth, trim=0cm 17.0cm 8.2cm 0cm, clip]{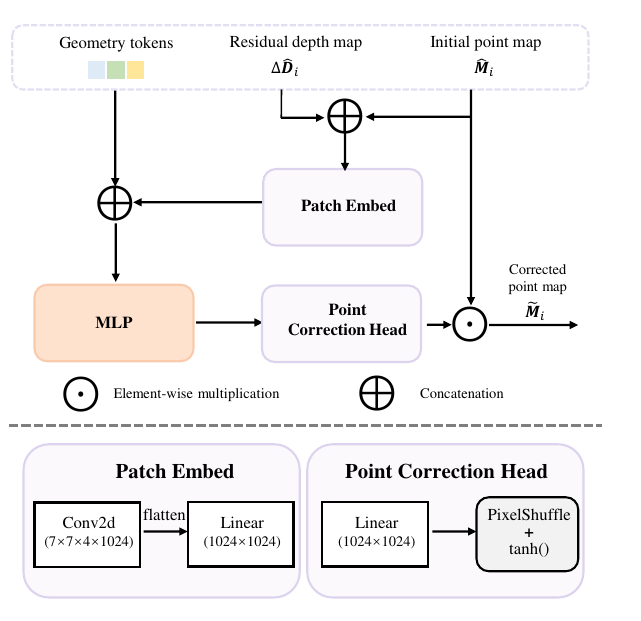}
\caption{\textbf{Residual-Guided Geometry Correction (RGC).}
RGC uses the residual depth map $\Delta \hat{\mathbf{D}}_i$ from FGPR to correct the initial local point map $\hat{\mathbf{M}}_i$. 
The residual depth map and the initial point map are embedded and fused with geometry tokens to predict a bounded multiplicative correction, producing the corrected point map $\tilde{\mathbf{M}}_i$. 
}
\label{fig:point refine}
\end{figure}

To further improve the coarse local geometry, we introduce a lightweight \emph{Residual-Guided Geometry Correction (RGC)} module, as shown in Fig.~\ref{fig:point refine}. RGC takes the initial local point map together with the depth residual from FGPR as input, so that local geometry correction is explicitly guided by residual geometric cues after pose refinement.

For each input view, the initial point map $\mathbf{\hat{M}}_{i} \in \mathbb{R}^{H \times W \times 3}$ is concatenated with the FGPR depth residual $\Delta \mathbf{\hat{D}}_i \in \mathbb{R}^{H \times W \times 1}$ and converted into patch tokens
$\mathbf{z}_i^{r} \in \mathbb{R}^{\frac{4HW}{p^2} \times D}$ through a patch embedding layer:
\begin{equation}
\mathbf{z}_i^{r} =
\mathrm{PatchEmbed}\!\left(
\mathrm{Concat}(\mathbf{\hat{M}}_{i}, \Delta \mathbf{\hat{D}}_i)
\right),
\end{equation}
where $p=14$ denotes the patch size and $D=1024$ is the token dimension. 
The residual tokens are then combined with the geometry tokens 
$\mathbf{z}_i^{g}$ from the geometry branch, which share the same spatial resolution. 
We concatenate them along the channel dimension and use an MLP to obtain the correction tokens:
\begin{equation}
\mathbf{z}_i^{\delta} =
\mathrm{MLP}\!\left(
\mathrm{Concat}_{c}(\mathbf{z}_i^{r}, \mathbf{z}_i^{g})
\right),
\end{equation}
where $\mathbf{z}_i^{\delta} \in \mathbb{R}^{\frac{4HW}{p^2} \times D}$.
The correction tokens are decoded into a dense point-wise correction field $\Delta \hat{\mathbf{M}}_i \in \mathbb{R}^{H \times W \times 3}$ through linear projection and pixel-shuffle upsampling:
\begin{equation}
\Delta \hat{\mathbf{M}}_i =
\operatorname{PS}\!\left(
\mathrm{Linear}\!\left(\mathbf{z}_i^{\delta}\right)
\right),
\end{equation}
where $\operatorname{PS}(\cdot)$ denotes pixel-shuffle.
Rather than directly predicting additive offsets, the module estimates a bounded multiplicative correction,
\begin{equation}
\mathbf{S}_i = 1 + \beta \tanh(\Delta \mathbf{\hat{M}}_i),
\end{equation}
which is applied element-wise to the initial point map:
\begin{equation}
\mathbf{\tilde{M}}_{i} =
\mathbf{\hat{M}}_{i} \odot \mathbf{S}_i,
\end{equation}
where $\odot$ denotes element-wise multiplication, and $\beta=0.2$ constrains the correction magnitude to avoid excessive corrections that could disrupt the coarse geometric structure.

This design preserves the coarse geometric prior while enabling spatially adaptive geometry correction. Compared with direct offset regression, the multiplicative formulation yields more stable updates and is less prone to over-correction. By conditioning the refinement on FGPR depth residuals, the refined point map becomes better aligned with the corrected camera poses, providing more reliable Gaussian centers for subsequent 3D Gaussian prediction.

\subsection{Dual-Path Training Strategy}
\label{sec:dual_path}
\begin{figure*}[t]
\centering
\includegraphics[width=\linewidth, trim=0cm 21.6cm 0cm 0cm, clip]{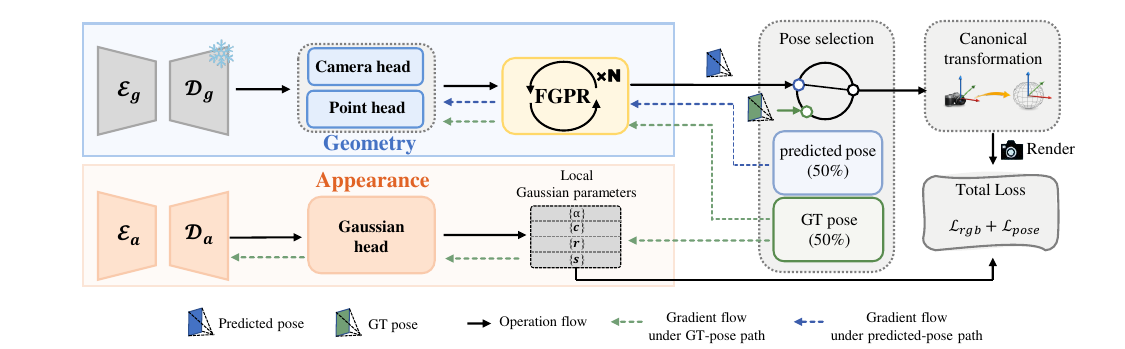}
\caption{\textbf{Dual-Path Training Strategy.}
The refined predicted pose and the ground-truth (GT) pose are sampled with equal probability. 
The selected pose is used to transform local Gaussian centers into the canonical space for rendering. 
When the ground-truth pose is selected, reliable rendering supervision updates both the geometry and attribute branches. 
When the predicted pose is selected, rendering gradients are detached from the attribute branch and only the optimize geometry branch. 
This asymmetric optimization stabilizes attribute learning while improving geometry robustness under residual pose errors.}
\label{fig:dual-branch train}
\end{figure*}
We adopt a dual-branch design for SG-Splat, where the geometry branch estimates camera poses and local Gaussian centers, and the attribute branch predicts Gaussian attributes. 
This separation reduces error propagation from inaccurate poses and improves the reliability of Gaussian parameter generation. 
However, 3DGS rendering requires transforming local Gaussian centers into a shared canonical space, making cross-view alignment sensitive to pose errors. 
To mitigate this issue, we introduce a mixed-pose canonical transformation with asymmetric optimization, as shown in Fig.~\ref{fig:dual-branch train}.

\paragraph{Mixed-pose canonical transformation}
The refined point map $\tilde{\mathbf{M}}_{i} \in \mathbb{R}^{H \times W \times 3}$ provides local Gaussian centers for the $i$-th input view. 
After flattening the spatial dimensions, we obtain
$\{\boldsymbol{\mu}_{i,j}^{\mathrm{local}}\}_{j=1}^{HW}$. 
During training, the pose used for canonical transformation is sampled from either the ground-truth pose or the refined predicted pose:
\begin{equation}
\bar{\mathbf{P}}_{i} =
\begin{cases}
\mathbf{P}_{i}^{*}, & z = 1,\\
\tilde{\mathbf{P}}_{i}, & z = 0,
\end{cases}
\qquad
z \sim \mathrm{Bernoulli}(0.5),
\label{eq:pose_selection}
\end{equation}
where $\mathbf{P}_{i}^{*}$ and $\tilde{\mathbf{P}}_{i}$ denote the ground-truth pose and refined predicted pose, respectively. 
The binary variable $z$ is sampled once per training scene and shared across all input views.
Each local center is transformed into the canonical space by
\begin{equation}
\boldsymbol{\mu}_{i,j}^{\mathrm{global}}
=
\bar{\mathbf{P}}_{i} \circ \boldsymbol{\mu}_{i,j}^{\mathrm{local}} .
\label{eq:canonical_transform_center}
\end{equation}

The attribute branch predicts the remaining Gaussian attributes for each center:
\begin{equation}
\left\{
(\mathbf{c}_{i,j}, \mathbf{r}_{i,j}, \mathbf{s}_{i,j}, \alpha_{i,j})
\right\}_{j=1}^{HW}.
\end{equation}
Together with the transformed centers
$\{\boldsymbol{\mu}_{i,j}^{\mathrm{global}}\}_{j=1}^{HW}$, these predicted parameters form the Gaussian primitives for rendering. 
For $N$ input views, the model generates $N \times H \times W$ primitives in total.

\paragraph{Asymmetric optimization under mixed poses}
The optimization rule depends on the sampled pose in Eq.~\eqref{eq:pose_selection}. 
When $z=1$, ground-truth poses provide reliable canonical transformation and clean rendering supervision. 
We therefore update the full model, including the geometry branch, pose refinement module, and attribute branch. 
This path stabilizes Gaussian attribute learning under well-aligned transformations.
When $z=0$, refined predicted poses are used, matching the inference setting where ground-truth poses are unavailable. 
However, residual pose errors may misalign Gaussian centers across views and introduce noisy rendering supervision. 
To avoid corrupting attribute learning, we detach the rendering gradients from the attribute branch and use them only to optimize the geometry branch.

This asymmetric design combines the advantages of both paths. 
The ground-truth-pose path provides reliable supervision for Gaussian attribute prediction, while the predicted-pose path exposes the geometry branch to the pose distribution used during inference. 
By alternating between them, the proposed strategy reduces the discrepancy between training and inference and prevents the attribute branch from absorbing pose-induced noise.

\subsection{Training Objectives}
\paragraph{Progressive Decoupled Training}
Directly integrating FGPR and RGC into the SG-Splat pipeline leads to unstable optimization, as pose refinement, geometry correction, and Gaussian attribute prediction are optimized simultaneously. The resulting entanglement makes it difficult for the refinement modules to learn reliable corrections from noisy pose and geometry estimates. To address this issue, we adopt a progressive training strategy that decouples geometry and appearance optimization across four stages, including coupled pretraining, pose refinement, point refinement, and final joint optimization.

\textbf{Stage I: Coupled pretraining.}
We first train a coupled base model without FGPR and RGC, including the encoder, decoder, point head, camera head, and Gaussian head. This stage provides a stable initialization for coarse geometry reconstruction and Gaussian attribute prediction.

\textbf{Stage II: Pose refinement via FGPR.}
Based on the pretrained coupled model, we duplicate the encoder and decoder to initialize separate geometry and appearance branches. The point and camera heads are assigned to the geometry branch, while the Gaussian head is assigned to the appearance branch. All other modules are frozen, and only FGPR is optimized to learn stable pose refinement.

\textbf{Stage III: Point refinement via RGC.}
We then freeze the remaining components and optimize only the RGC, enabling geometry correction under the refined poses without interference from appearance optimization. During the refinement stages, local points are transformed into the canonical space using ground-truth poses, which removes stochastic perturbations from inaccurate pose predictions and provides stable supervision for pose and geometry refinement.

\textbf{Stage IV: Joint optimization.}
Finally, we fine-tune the full dual-path architecture in an end-to-end manner. With stabilized geometry and pose estimates as initialization, this stage jointly optimizes geometry and appearance, improving reconstruction fidelity, view consistency, and robustness to residual pose errors.
\begin{table*}[t]
\centering
\caption{Quantitative comparisons on DL3DV~\cite{ling2024dl3dv} under different input-view settings. 
GT Pose indicates whether camera poses are provided. 
Our method consistently achieves the best performance from 12 to 32 input views in both posed and pose-free settings.}
\label{tab:main_result}
\scriptsize
\setlength{\tabcolsep}{3.5pt}
\renewcommand{\arraystretch}{1.08}
\resizebox{\textwidth}{!} {
\begin{tabular}{lc|ccc|ccc|ccc|ccc}
\toprule
\multirow{2}{*}{Method} & \multirow{2}{*}{GT Pose}
& \multicolumn{3}{c|}{12v}
& \multicolumn{3}{c|}{18v}
& \multicolumn{3}{c|}{24v}
& \multicolumn{3}{c}{32v} \\
\cmidrule(lr){3-5} \cmidrule(lr){6-8} \cmidrule(lr){9-11} \cmidrule(lr){12-14}
& 
& PSNR$\uparrow$ & SSIM$\uparrow$ & LPIPS$\downarrow$
& PSNR$\uparrow$ & SSIM$\uparrow$ & LPIPS$\downarrow$
& PSNR$\uparrow$ & SSIM$\uparrow$ & LPIPS$\downarrow$
& PSNR$\uparrow$ & SSIM$\uparrow$ & LPIPS$\downarrow$ \\
\midrule
MVSplat      & \checkmark & 21.149 & 0.696 & 0.236 & 19.452 & 0.632 & 0.281 & 19.761 & 0.641 & 0.272 & 17.912 & 0.492 & 0.403 \\
DepthSplat   & \checkmark & 22.027 & 0.763 & 0.206 & 19.712 & 0.673 & 0.294 & 20.110 & 0.695 & 0.275 & 18.240 & 0.601 & 0.365 \\
YoNoSplat    & \checkmark & \underline{23.285} & \underline{0.773} & \underline{0.177} & \underline{22.428} & \underline{0.747} & \underline{0.199} & \underline{22.664} & \underline{0.758} & \underline{0.192} & \underline{21.728} & \underline{0.727} & \underline{0.217} \\
\rowcolor{gray!15}
Ours         & \checkmark & \textbf{23.487} & \textbf{0.781} & \textbf{0.168} & \textbf{22.615} & \textbf{0.764} & \textbf{0.179} & \textbf{23.163} & \textbf{0.775} & \textbf{0.174} & \textbf{22.255} & \textbf{0.745} & \textbf{0.187} \\
\midrule
NoPoSplat    &            & 19.182 & 0.543 & 0.338 & 17.325 & 0.501 & 0.401  & 17.641 & 0.498 & 0.399 & 17.024 & 0.481 & 0.410 \\
AnySplat     &            & 19.493 & 0.608 & 0.314 & 19.241 & 0.614 & 0.324  & 19.914 & 0.635 & 0.312 & 19.503 & 0.620 & 0.325 \\
WorldMirror &             & 21.629 & 0.673 & 0.184 & 20.190 & 0.684 & 0.248  & 20.664 & 0.697 & 0.241 & 20.369 & 0.689 & 0.248 \\
YoNoSplat    &            & \underline{23.151} & \underline{0.759} & \underline{0.182} & \underline{22.334} & \underline{0.733} & \underline{0.207}  & \underline{22.433} & \underline{0.735} & \underline{0.205} & \underline{21.525} & \underline{0.702} & \underline{0.231} \\
\rowcolor{gray!15}
Ours         &            & \textbf{23.791} & \textbf{0.790} & \textbf{0.156} & \textbf{22.982} & \textbf{0.764} & \textbf{0.178} &  \textbf{23.190} & \textbf{0.772} & \textbf{0.171} & \textbf{22.430} & \textbf{0.744} & \textbf{0.193} \\
\bottomrule
\end{tabular}
}
\end{table*}

\paragraph{Training Loss}
Our model is a joint learning framework and trained with a multi-task loss function comprising two components:
camera pose loss, and 3DGS rendering loss.
As for the pose loss, we only explicitly supervise the refined poses produced by the FGPR module. 
Given the ground-truth poses $\{\mathbf{P}_i^*\}_{i=1}^{N}$, we supervise the refined pose estimates 
$\{\tilde{\mathbf{P}}_i^{(t)}\}_{i=1}^{N}$ over $T$ iterations by
\begin{equation}
\mathcal{L}_{\mathrm{pose}}
=
\sum_{t=1}^{T} \gamma^{T-t}
\sum_{i=1}^{N}
\left\|
\log\!\left(
{\mathbf{P}_i^*}^{-1} \tilde{\mathbf{P}}_i^{(t)}
\right)
\right\|_2 ,
\end{equation}
where $\log(\cdot)$ denotes the $\mathrm{SE}(3)$ logarithm map and $\gamma=0.9$ emphasizes later refinement iterations.
We do not impose direct depth supervision because 3DGS renders both images and depths via $\alpha$-blending, such that accurate pixel-level depth does not necessarily imply accurate rendered depth. In addition, the predicted depths are defined in the camera coordinate system, making it difficult to enforce cross-view consistency. 

The rendering loss is defined as the sum of the MSE and LPIPS~\cite{zhang2018lpips} losses over all target views rendered in a single forward pass:
\begin{equation}
\mathcal{L}_{\mathrm{rgb}}
=
\sum_{n=1}^{N_{\mathrm{tgt}}}
\left(
\mathrm{MSE}(\tilde{I}^{\,n}, I_{\mathrm{gt}}^{\,n})
+
\lambda \,\mathrm{LPIPS}(\tilde{I}^{\,n}, I_{\mathrm{gt}}^{\,n})
\right),
\end{equation}
where $N_{\mathrm{tgt}}$ denotes the number of target views in each forward pass, $\tilde{I}^{\,n}$ is the rendered image, and $I_{\mathrm{gt}}^{\,n}$ is the corresponding ground-truth image. The overall training objective is then defined as
\begin{equation}
\mathcal{L}
=
\mathcal{L}_{\mathrm{rgb}}
+
\beta\,\mathcal{L}_{\mathrm{pose}},
\end{equation}
where $\beta$ controls the relative importance of pose supervision.

\section{Experiments}

\begin{figure*}[t]
\centering
\includegraphics[width=\linewidth, trim=0cm 13.4cm 0cm 0cm, clip]{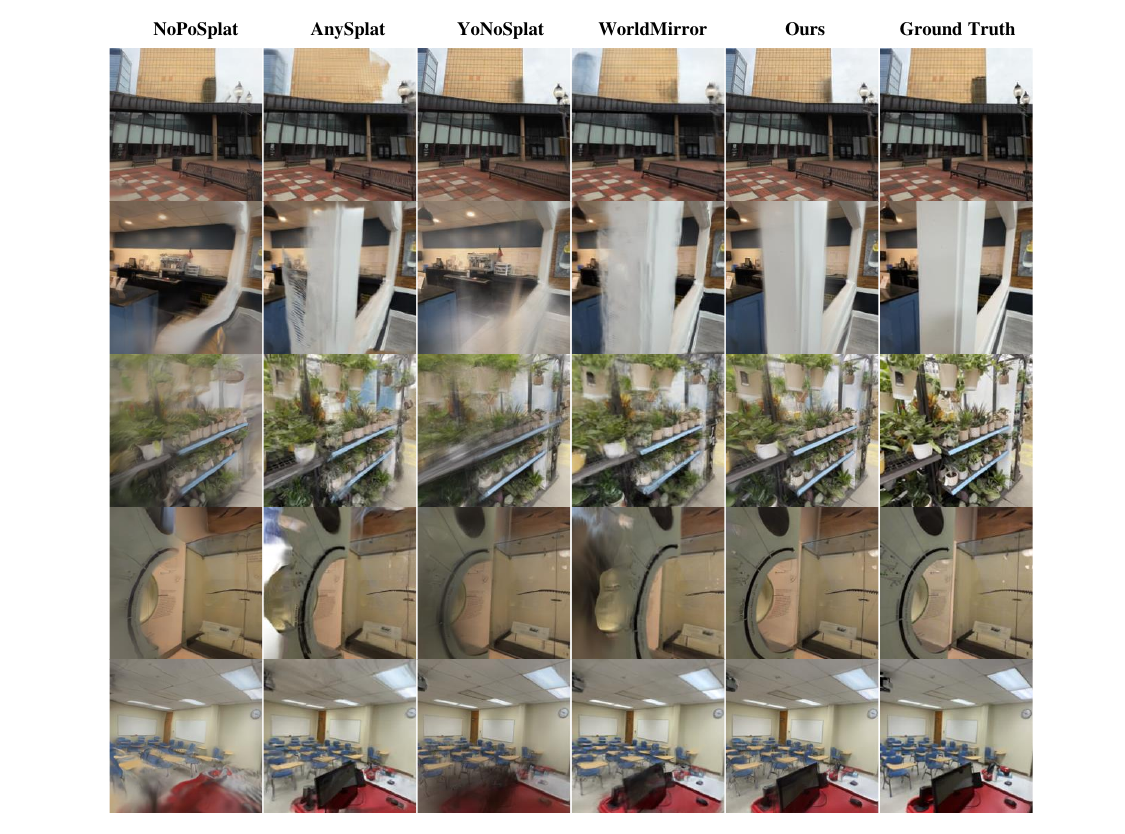}
\caption{\textbf{Qualitative comparisons on the DL3DV dataset with 24 input views.} With more accurate camera pose estimation, our model produces geometrically consistent Gaussian primitives, yielding sharper boundaries, finer details, and fewer artifacts in novel views.}
\label{fig:comparison_img_dl3dv}
\end{figure*}

\begin{table}[t]
\centering
\caption{Quantitative comparisons on RealEstate10K~\cite{zhou2018re10k} under the 6-view setting. 
GT Pose indicates whether camera poses are provided.}
\label{tab:quantity_re10k}
\begin{tabular}{lcccc}
\toprule
Method & GT Pose & PSNR$\uparrow$ & SSIM$\uparrow$ & LPIPS$\downarrow$ \\
\midrule
MVSplat & \checkmark & 22.456 & 0.761 & 0.223 \\
DepthSplat & \checkmark & 24.172 & 0.848 & 0.141 \\
\rowcolor{gray!15}
Ours       & \checkmark & \textbf{25.041} & \textbf{0.849} & \textbf{0.132} \\
\midrule
NoPoSplat  &           & 22.044 & 0.731 & 0.210 \\
AnySplat  &           & 23.712 & 0.748 & 0.186 \\
WorldMirror &           & 23.795 & 0.844 & 0.152 \\
YoNoSplat  &           & 25.334 & 0.848 & 0.135 \\
\rowcolor{gray!15}
Ours       &           & \textbf{25.478} & \textbf{0.863} & \textbf{0.130} \\

\bottomrule
\end{tabular}
\end{table}


\begin{figure*}[t]
\centering
\includegraphics[width=\linewidth, trim=0cm 17.4cm 0cm 0cm, clip]{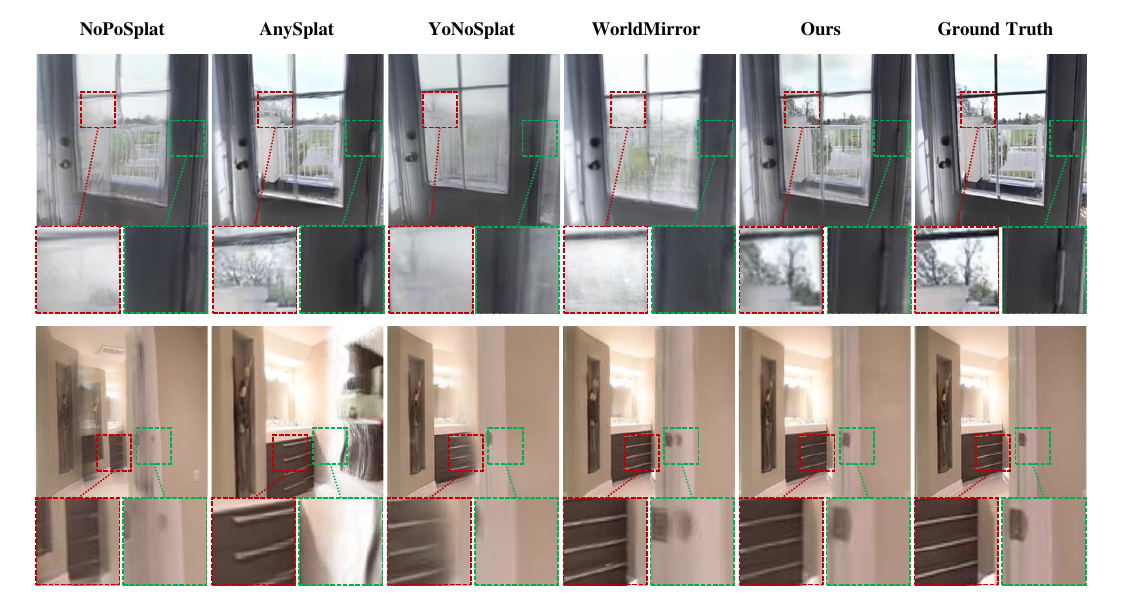}
\caption{\textbf{Qualitative comparisons on the RealEstate10K dataset with sparse-view inputs (6 views).} Even with only a few input views, our model reconstructs more coherent scene geometry and more realistic novel views, with clearer details and reduced artifacts.}
\label{fig:comparison_img_re10k}
\end{figure*}
\subsection{Experimental Setup}
\paragraph{Implementation Details}
For both geometry and appearance branches, the encoder employs the DINOv2 Large model with 24 attention layers, and the decoder consists of
18 alternating-attention layers. The backbones, encoders, decoders, point head, camera head, and Gaussian head are initialized from the pretrained YoNoSplat~\cite{ye2025yonosplat} weights. 
For FGPR, the feature encoder, context encoder, and update operator are initialized from the pretrained DROID-SLAM model~\cite{teed2021droid}. 
The newly introduced layers are initialized to zero, so that they do not perturb the pretrained representations at the beginning of training.
During training, we randomly select the number of input views between 2 and 32 and sample 4 target views. All experiments are conducted at a resolution of $224 \times 224$. We train the model on a single NVIDIA L20 GPU with a batch size of 1 for 100k steps. In the FGPR module, the number of pose refinement iterations $T$ is set to 2. For the training objective, we set $\lambda=0.05$ for the LPIPS term and $\beta=0.02$ for the pose supervision term.
\paragraph{Datasets}
We train and evaluate our model on the DL3DV~\cite{ling2024dl3dv} and RealEstate10K~\cite{zhou2018re10k} datasets.
DL3DV contains 51.3 million frames captured from 10,510 real-world scenes, spanning 65 point-of-interest categories. We use the full training set for model training, while excluding the 140 benchmark scenes to avoid potential evaluation leakage. We evaluate novel view synthesis under different input-view settings, including 12, 18, 24, and 32 views, with 8 target views used for evaluation.
RealEstate10K consists of real-estate videos collected from YouTube, where diverse views of a scene are obtained by sampling frames at different timestamps. We use this dataset to evaluate sparse-view reconstruction with 6 input views.
\paragraph{Evaluation Metrics}
We evaluate novel view synthesis using the widely adopted metrics PSNR, SSIM\cite{wang2004ssim}, and LPIPS~\cite{zhang2018lpips}. For pose estimation, we follow prior works~\cite{jiang2025anysplat,ye2025yonosplat} and report the area under the cumulative angular pose error curve (AUC) at angular thresholds of $5^\circ$, $10^\circ$, and $20^\circ$.

\paragraph{Baselines}
We compare our method with representative state-of-the-art feed-forward 3DGS methods under both pose-free and pose-dependent settings. For pose-free reconstruction, we consider the following methods:
\begin{itemize}
    \item \textbf{YoNoSplat}~\cite{ye2025yonosplat} is a unified feed-forward 3DGS model designed for both pose-dependent and pose-free reconstruction. It predicts per-view Gaussian primitives and camera poses and aggregates them into a global representation using the predicted camera poses. In our experiments, we use the publicly released checkpoints trained on DL3DV and RealEstate10K at $224 \times 224$ resolution, respectively.
    
    \item \textbf{NoPoSplat}~\cite{noposplat} reconstructs 3D Gaussians from sparse unposed images by anchoring one input view as the canonical coordinate system, thereby avoiding explicit local-to-global Gaussian transformation with estimated poses. For a fair comparison, we retrain NoPoSplat under the corresponding multi-view settings on each dataset for 150K steps. On RealEstate10K, we initialize the model from the official pretrained 3-view weights, while keeping the remaining training configurations consistent with the original paper.
    
    \item \textbf{AnySplat}~\cite{jiang2025anysplat} targets feed-forward 3DGS from unconstrained image collections. It jointly predicts Gaussian parameters, depth maps, and camera poses from uncalibrated inputs, with geometry supervision provided by pretrained VGGT priors.
    
    \item \textbf{WorldMirror}~\cite{liu2025worldmirror} is a general-purpose feed-forward 3D geometric prediction model that can incorporate different geometric priors and output multiple 3D representations, including camera parameters, depth maps, point clouds, and 3D Gaussians.  We use its publicly released checkpoint trained on multiple large-scale datasets, including DL3DV and RealEstate10K.
\end{itemize}

We further compare with pose-dependent feed-forward 3DGS methods, where ground-truth camera poses are provided as input:
\begin{itemize}
    \item \textbf{MVSplat}~\cite{mvsplat} constructs plane-sweeping cost volumes from posed sparse-view images and uses the estimated depth maps to localize Gaussian centers. For RealEstate10K, we evaluate the official checkpoint released by the authors. For DL3DV, no official checkpoint is available for our multi-view setting. Therefore, we retrain MVSplat for 300K iterations following the original training protocol and extend the number of input views to 32 views.
    
    \item \textbf{DepthSplat}~\cite{xu2024depthsplat} enhances cost-volume depth estimation with pretrained monocular depth features, leading to more robust Gaussian center prediction under challenging matching conditions. We use the official checkpoints trained on DL3DV and RealEstate10K.
\end{itemize}

For camera pose estimation, we additionally report results from recent feed-forward 3D Foundation Models, including \textbf{MASt3R}~\cite{mast3r}, \textbf{VGGT}~\cite{wang2025vggt} and $\boldsymbol{\pi^3}$~\cite{wang2025pi3}.

\begin{table*}[t]
\centering
\caption{\textbf{Pose estimation results on DL3DV and RealEstate10K.} 
The upper block reports 3D foundation models, and the lower block reports pose-free feed-forward 3DGS methods. 
Our method achieves the best AUC across dense-view DL3DV settings and sparse-view RealEstate10K.}
\label{tab:pose_est}
\small
\setlength{\tabcolsep}{5pt}
\renewcommand{\arraystretch}{1.55}

\begin{tabular}{lccccccccccccccc}
\toprule
\multirow{3}{*}{Method} 
& \multicolumn{12}{c}{DL3DV} 
& \multicolumn{3}{c}{RealEstate10K} \\

\cmidrule(lr){2-13} \cmidrule(lr){14-16}

& \multicolumn{3}{c}{12v}
& \multicolumn{3}{c}{18v} 
& \multicolumn{3}{c}{24v} 
& \multicolumn{3}{c}{32v} 
& \multicolumn{3}{c}{6v} \\

\cmidrule(lr){2-4} \cmidrule(lr){5-7} \cmidrule(lr){8-10} \cmidrule(lr){11-13} \cmidrule(lr){14-16}

& $5^\circ$ & $10^\circ$ & $20^\circ$
& $5^\circ$ & $10^\circ$ & $20^\circ$
& $5^\circ$ & $10^\circ$ & $20^\circ$
& $5^\circ$ & $10^\circ$ & $20^\circ$
& $5^\circ$ & $10^\circ$ & $20^\circ$ \\

\midrule
MASt3R              
& 0.626 & 0.773 & 0.874 
& 0.599 & 0.756 & 0.866 
& 0.638 & 0.784 & 0.883 
& 0.653 & 0.788 & 0.881 
& 0.521 & 0.727 & 0.770 \\
VGGT              
& 0.680 & 0.838 & 0.919 
& 0.654 & 0.821 & 0.908
& 0.653 & 0.824 & 0.911 
& 0.629 & 0.809 & 0.903 
& 0.590 & 0.763 & 0.870 \\

${\pi^3}$         
& 0.837 & 0.918 & 0.959 
& \textbf{0.792} & \underline{0.895} & \underline{0.947} 
& \underline{0.829} & \underline{0.914} & \underline{0.957} 
& \underline{0.797} & \underline{0.899} & \underline{0.949} 
& 0.725 & 0.852 & 0.921 \\

\midrule
NoPoSplat         
& 0.412 & 0.625 & 0.744 
& 0.393 & 0.612 & 0.735 
& 0.401 & 0.635 & 0.751 
& 0.390 & 0.625 & 0.739 
& 0.431 & 0.619 & 0.749 \\

AnySplat          
& 0.606 & 0.794 & 0.896
& 0.576 & 0.768 & 0.879 
& 0.593 & 0.782 & 0.888 
& 0.570 & 0.770 & 0.880 
& 0.512 & 0.700 & 0.830 \\

WorldMirror       
& \underline{0.840} & \underline{0.920} & \underline{0.960} 
& \underline{0.774} & 0.885 & 0.941 
& 0.792 & 0.895 & 0.947 
& 0.772 & 0.883 & 0.941 
& \underline{0.762} & 0.871 & 0.931 \\

YoNoSplat         
& 0.833 & 0.917 & 0.958 
& 0.743 & 0.870 & 0.935 
& 0.799 & 0.895 & 0.947 
& 0.759 & 0.875 & 0.937 
& 0.745 & \underline{0.878} & \underline{0.932} \\

\rowcolor{gray!15}
Ours 
& \textbf{0.868} & \textbf{0.934} & \textbf{0.967} 
& \textbf{0.792} & \textbf{0.896} & \textbf{0.948} 
& \textbf{0.849} & \textbf{0.920} & \textbf{0.960} 
& \textbf{0.809} & \textbf{0.901} & \textbf{0.950} 
& \textbf{0.801} & \textbf{0.896} & \textbf{0.946} \\

\bottomrule
\end{tabular}
\end{table*}

\begin{table}[t]
\centering
\caption{Ablation study on DL3DV with 24 input views. FGPR* indicates that FGPR is trained, with all others frozen. The base model adopts the same architecture but excludes both FGPR and RGC, and is trained with ground-truth (GT) poses.}
\label{tab:ablation_main}
\small
\setlength{\tabcolsep}{6pt}
\begin{tabular}{lccc}
\toprule
Variant & PSNR$\uparrow$ & SSIM$\uparrow$ & LPIPS$\downarrow$ \\
\midrule
Base               & 22.361 & 0.735 & 0.203 \\
FGPR*        & 22.499 & 0.740 & 0.201 \\
FGPR + Pred Pose  & 21.155 & 0.647 & 0.213\\
FGPR + GT Pose  & 22.912 & 0.751 & 0.192\\
FGPR + Mix Pose  & 23.078 & 0.763 & 0.181\\
FGPR + Mix Pose + RGC  & \textbf{23.190} & \textbf{0.772} & \textbf{0.171}\\
\bottomrule
\end{tabular}
\end{table}
\begin{table}[t]
\centering
\caption{Effect of the number of FGPR iterations $T$ on DL3DV with 24 input views.}
\label{tab:ablation_iter}
\small
\setlength{\tabcolsep}{8pt}
\begin{tabular}{ccccc}
\toprule
$T$ & PSNR$\uparrow$ & SSIM$\uparrow$ & LPIPS$\downarrow$ & AUC@5$^\circ\uparrow$ \\
\midrule
1 & 23.079 & 0.761 & 0.179 & 0.794 \\
2 & \textbf{23.190} & \textbf{0.772} & 0.171 & 0.849 \\
3 & 23.183 & 0.770 & \textbf{0.170} & \textbf{0.851} \\
\bottomrule
\end{tabular}
\end{table}

\subsection{Novel View Synthesis}

Tab.~\ref{tab:main_result} and Tab.~\ref{tab:quantity_re10k} show the quantitative results for novel view synthesis on DL3DV and RealEstate10K. 
The symbol $p$ indicates whether ground-truth camera poses are provided as input. 
On DL3DV, our method achieves the best results across all evaluated view settings in both pose-dependent and pose-free configurations. 
In the posed setting, SG-Splat achieves the best quantitative results across all view settings.
This demonstrates that the proposed framework remains effective even when accurate camera poses are available.
In the pose-free setting, SG-Splat also achieves the best quantitative results across all view settings, with a maximum PSNR gain of 0.91 dB over the second-best result.
Although the absolute difficulty varies across view settings, the relative advantage of our method becomes more pronounced under denser input settings. 
This property is particularly important for feed-forward 3DGS, where dense inputs introduce more view-dependent Gaussian primitives, making accurate geometric correction crucial for mitigating pose-induced errors and ensuring reliable Gaussian reconstruction.

On RealEstate10K, SG-Splat further demonstrates its effectiveness under the pose-free sparse-view setting. 
It achieves the best overall quantitative performance across the evaluation metrics, outperforming all baselines. 
This confirms that our method remains robust in sparse real-world scenes, rather than being limited to dense-view reconstruction.

Qualitative comparisons are shown in Fig.~\ref{fig:comparison_img_dl3dv} and Fig.~\ref{fig:comparison_img_re10k}. 
Our method produces sharper textures, cleaner object boundaries, and more complete scene structures. 
In contrast, AnySplat tends to generate blurred details because inaccurate predicted poses perturb the transformed Gaussian centers and degrade the supervision of Gaussian attributes. 
Other methods often suffer from distorted geometry or missing structures due to inaccurate pose estimation and geometric reconstruction.
With the proposed FGPR module, our method improves pose accuracy before Gaussian reconstruction, leading to more reliable geometric alignment and more stable Gaussian parameter prediction. 
As a result, the rendered novel views are visually sharper and geometrically more consistent.

\subsection{Relative Pose Estimation}

We further evaluate the relative pose estimation on DL3DV and RealEstate10K, as reported in Tab.~\ref{tab:pose_est}. 
DL3DV is evaluated under dense-view settings with 12, 18, 24, and 32 input views, while RealEstate10K is evaluated under the sparse-view setting with 6 input views. 
The comparison includes pose-free feed-forward 3DGS methods and 3D foundation models, including MASt3R, VGGT, and $\pi^3$.

Our method achieves the best or comparable-best pose estimation accuracy across the evaluated settings. 
Our method achieves the highest AUC values on DL3DV under most evaluated view settings and consistently outperforms the compared methods on RealEstate10K across all AUC thresholds.
The consistent performance across dense-view DL3DV and sparse-view RealEstate10K demonstrates the robustness of the proposed pose estimation framework.

The comparison with $\pi^3$ and VGGT is particularly informative. 
These models are trained on large-scale multi-dataset corpora that include DL3DV and RealEstate10K, whereas our method is designed and trained for the feed-forward 3DGS reconstruction task. 
Despite this strong comparison setting, our method still achieves superior or comparable pose accuracy on most evaluated protocols. 
These results indicate that the proposed FGPR module effectively improves both fine-grained pose accuracy and coarse relative pose estimation, which further contributes to the superior rendering quality reported in Tab.~\ref{tab:main_result} and Tab.~\ref{tab:quantity_re10k}.

\subsection{Ablation Study}
We conduct ablation experiments on DL3DV under the 24-view setting. The base model adopts the same architecture but excludes the proposed Flow-Guided Pose Refinement (FGPR) and Residual-Guided Geometry Correction (RGC) modules, and is trained with ground-truth (GT) poses. We progressively incorporate FGPR, different pose selection strategies, and RGC to evaluate their individual contributions to rendering performance.

\paragraph{Effect of FGPR}
As shown in Tab.~\ref{tab:ablation_main}, introducing FGPR alone (FGPR*) leads to only marginal improvements over the base model. This suggests that while pose refinement is beneficial, its effectiveness is limited when optimized in isolation with all other components frozen, and cannot fully compensate for pose inaccuracies during training.

\paragraph{Effect of Dual-Branch Pose Selection Strategy}
We further investigate different pose selection strategies for training FGPR. Training with predicted poses (Pred Pose) leads to a noticeable performance degradation, indicating that noisy pose estimates introduce instability and hinder effective optimization. In contrast, using only GT poses achieves strong performance but suffers from exposure bias, as the model is not exposed to pose perturbations during training. Our proposed dual-branch strategy (Mix Pose), which adaptively combines predicted and GT poses, achieves the best results. This demonstrates that jointly leveraging accurate poses and realistic pose perturbations effectively reduces the discrepancy between training and inference, resulting in more robust rendering quality.

\paragraph{Effect of RGC}
Building upon FGPR with the proposed dual-path training strategy, further incorporating RGC consistently improves all metrics. This suggests that point-map correction driven by depth residuals improves the reliability of Gaussian centers, thereby enhancing rendering fidelity.

\paragraph{Effect of the Number of Iterations}
Tab.~\ref{tab:ablation_iter} evaluates the effect of the number of FGPR iterations $T$. A single iteration ($T=1$) yields suboptimal performance, indicating that one-step refinement is insufficient to fully correct pose errors. 
Increasing $T$ to 2 significantly improves all metrics, demonstrating that iterative refinement is effective in enhancing pose quality and rendering accuracy. 
Increasing $T$ from 2 to 3 slightly improves LPIPS and AUC@5$^\circ$, but reduces PSNR and SSIM while increasing computational cost. 
We therefore set $T=2$ as a practical trade-off between reconstruction quality, pose accuracy, and efficiency.

\section{Conclusion}
In this paper, we presented SG-Splat, a pose-free feed-forward 3D Gaussian Splatting framework for robust reconstruction under inaccurate pose estimation. 
To reduce the impact of residual pose errors, SG-Splat decouples geometry estimation from Gaussian attribute prediction and introduces a coarse-to-fine refinement pipeline with Flow-Guided Pose Refinement and Residual-Guided Geometry Correction. 
We further proposed a dual-path training strategy that mixes ground-truth and predicted poses with asymmetric optimization, improving robustness to pose errors while maintaining reliable pose supervision. 
Extensive experiments on DL3DV and RealEstate10K show that SG-Splat achieves improved novel view synthesis and relative pose estimation under both sparse and dense input-view settings.

Despite these improvements, several limitations remain. 
First, our method assumes known camera intrinsics, limiting its applicability to in-the-wild images with unavailable or unreliable intrinsics. 
Second, pixel-aligned feed-forward reconstruction tends to produce view-dependent and redundant Gaussian primitives, resulting in less compact representations. 
Third, high-resolution reconstruction still incurs substantial training and inference costs. 
Future work will explore intrinsic-free reconstruction, more efficient high-resolution architectures, and non-pixel-aligned Gaussian prediction to obtain more compact and effective 3DGS representations.

\bibliographystyle{IEEEtran}
\bibliography{reference}

\end{document}